\documentclass[runningheads]{llncs}

\usepackage{graphicx}
\usepackage{amsmath,amssymb}
\usepackage{mathtools}
\usepackage{xspace}
\usepackage{booktabs} 
\usepackage{xcolor}
\usepackage{rotating}
\usepackage{colortbl}
\usepackage{multirow}
\usepackage{array}
\usepackage{hyperref}
\usepackage{url}
\usepackage{caption}
\usepackage{subcaption}
\newcommand{\al}{\textit{et al.}}

\renewcommand{\epsilon}{\varepsilon}
\newcommand{\R}{\mathbb{R}}

\newcommand{\BBOB}{BBOB }
\renewcommand{\phi}{\varphi}

\begin{document}

\title{Towards Explainable Exploratory Landscape Analysis: Extreme Feature Selection for Classifying BBOB Functions}
\titlerunning{Towards Explainable ELA: Feature Selection for BBOB Functions}

\author{
Quentin Renau\inst{1,2} \and
Johann Dreo\inst{1}\and
Carola Doerr\inst{3}
\and
Benjamin Doerr\inst{2}
}
\authorrunning{Q. Renau, J. Dreo, C. Doerr, B. Doerr}
\institute{
Thales Research \& Technology, Palaiseau, France\\ \and
\'Ecole Polytechnique, Institut Polytechnique de Paris, CNRS, LIX, 
France \\ \and
Sorbonne Universit\'e, CNRS, LIP6, Paris, France\\ 
}

\maketitle              
\begin{abstract}
Facilitated by the recent advances of Machine Learning~(ML), the automated design of optimization heuristics is currently shaking up evolutionary computation~(EC). Where the design of hand-picked guidelines for choosing a most suitable heuristic has long dominated research activities in the field, automatically trained heuristics are now seen to outperform human-derived choices even for well-researched optimization tasks. ML-based EC is therefore not any more a futuristic vision, but has become an integral part of our community. 

A key criticism that ML-based heuristics are often faced with is their potential lack of explainability, which may hinder future developments. This applies in particular to supervised learning techniques which extrapolate algorithms' performance based on exploratory landscape analysis~(ELA). In such applications, it is not uncommon to use dozens of problem features to build the models underlying the specific algorithm selection or configuration task. Our goal in this work is to analyze whether this many features are indeed needed. Using the classification of the BBOB test functions as testbed, we show that a surprisingly small number of features -- often less than four -- can suffice to achieve a 98\% accuracy. Interestingly, the number of features required to meet this threshold is found to decrease with the problem dimension. We show that the classification accuracy transfers to settings in which several instances are involved in training and testing. In the leave-one-instance-out setting, however, classification accuracy drops significantly, and the transformation-invariance of the features becomes a decisive success factor.  

\keywords{Exploratory Landscape Analysis \and Feature Selection \and Black-Box Optimization.}
\end{abstract}
\sloppy{

\section{Introduction} \label{sec:intro}
Evolutionary algorithms and other iterative optimization heuristics (IOHs) are classically introduced as frameworks within which a user can gather some modules to instantiate an algorithm.
For instance, the {\em design} of an evolutionary algorithm requires to choose the population size, the variation and selection operators in use, the encoding structure, fitness function penalization weights, etc. 
This highly flexible design of IOHs allows for efficient abstractions but comes at the burden of having to solve an additional (meta-)optimization problem.
\emph{Automated design} of heuristics aims at solving this problem by providing data-driven recommendations which IOH shall be employed for a given optimization problem and how it shall be configured. Automated IOH design has proven its promise in numerous applications, see~\cite{HutterKV19,kerschke_automated_2019,MunozSurvey15,BelkhirDSS17,KerschkeT19} for examples and further references.  

A common critique of machine-trained automated algorithm design is its potential lack of explainability. That is, the general fear is that by relying on automated design approaches, we may be loosing intuition for \emph{why} certain recommendation are made -- a key driver for the development of new optimization approaches. This fear is not without any reason: the vast majority of automated algorithm design studies fall short in this explainability aspect. 

\textbf{Our Contribution.} 
Our work aims at providing paths to narrowing this important gap, by studying which information the trained models actually need to achieve convincing performance. As testbed we chose the automated classification of optimization problems through exploratory landscape analysis (ELA). We show that very small feature sets can suffice to reliably discriminate between various optimization problems and that these sets are robust with respect to the classifiers and function instances.

Apart from the explainability aspect, our findings have important consequences also for the efficiency of automated algorithm design: smaller feature sets are faster to compute and they can drastically reduce the time spent in the training phase. Another advantage of feature selection is that the classification or regression accuracy can \emph{increase}.

\textbf{Background and Motivation.} 
ELA was introduced in~\cite{mersmann_exploratory_2011} with the objective to gain insights about the properties of an unknown optimization problem. Instead of relying on expert knowledge, the keystone of ELA are computer-generated \emph{features} that are based on sampling the decision space.  
With the purpose of enhancing the effectiveness of this approach, several additional features have been introduced since. A good selection of these features are automatically computed by the R package \emph{flacco}~\cite{flacco2019}, see Sec.~\ref{sec:setup} for more details.

We chose classification as task, because it offers a very clean setting in which the results are easily interpretable. Classification has a straightforward performance measure, the \emph{classification accuracy}, i.e., the fraction of items that are classified correctly. 
Additionally, the classification accuracy is a good way of estimating the expressiveness of ELA feature sets, i.e., their ability to discriminate between different problems~\cite{renau_expressiveness_2019}.
A proper classification furthermore plays an important role also in many other ML tasks, including the selection and configuration of algorithms, so that a good classification accuracy can be expected to provide good results also for these tasks.

\textbf{Related Work.} Given the mentioned speed-up and the better performance  that one can expect from smaller feature sets, feature selection is not new, but rather standard in automated algorithm design. However, most related works still use a relatively large number of features, hindering explainability of the trained models. Among the ELA-based applications in EC, the following ones have used the smallest feature portfolios.  

Mu\~{n}oz and Smith-Miles~\cite{MunozS15dimension} compute the co-linearity between landscape features with the idea that if two features are strongly co-linear, they carry the same type of information about the landscape. 
Applying this procedure, nine features were kept for further analysis: the adjusted coefficient of determination of a linear regression model including interactions~\cite{mersmann_exploratory_2011}, the adjusted coefficient of determination of a quadratic regression model~\cite{mersmann_exploratory_2011}, the ratio between the minimum and maximum absolute values of the quadratic term coefficients in the quadratic model, the significance of $D$-th and first order~\cite{SeoM07}, the skewness, kurtosis and entropy of the fitness function distribution~\cite{mersmann_exploratory_2011}, and the maximum information content~\cite{munoz_exploratory_2015}.

Another method to perform feature selection is the use of search algorithms. In their work, Kerschke and Trautmann~\cite{KerschkeT19} compare four different algorithms, a greedy forward-backward selection, a greedy backward-forward selection, a $(10+5)$-GA and a $(10+50)$-GA. 
The smallest feature sets considered in their algorithm selection setting have a size of eight features: three features from the $y$-distribution feature set~\cite{mersmann_exploratory_2011} (skewness, kurtosis, and number of peaks), one level set feature~\cite{mersmann_exploratory_2011} (the ratio of mean misclassification errors when using a linear (LDA) and mixed discriminant analysis (MDA)), two information content features~\cite{munoz_exploratory_2015} (the maximum information content and the settling sensitivity), one cell mapping feature~\cite{ELA_cell} (the standard deviation of the distances between each cell’s center and worst observation), and one of the basic features (the best fitness value within the sample).
This result is still considerably larger than the sets we will identify as promising in our work. 

Saini \al~\cite{SainiLM19} and Lacroix and McCall~\cite{DBLP:conf/gecco/LacroixM19} also use reduced feature sets, but do not expand on how these have been derived.

\textbf{Availability of Our Data.} All our project data is available at~\cite{data}.

\section{Problem Classification via Majority Judgment} 
\label{sec:setup}
Our primary objective is to analyze the number of features that are needed to correctly classify the 24 BBOB functions from the COCO benchmark environment and their robustness across several dimensions and sample sizes. We describe in this section the benchmark set, the experimental procedure, and the classification scheme.

\textbf{The 24 BBOB Benchmark Problems.} 
A standard benchmark environment for numerical black-box optimization is the COCO (\textbf{CO}mparing \textbf{C}ontinuous \textbf{O}ptimizers) platform~\cite{cocoplat}.
From this environment, we consider the BBOB suite, a set of 24 noiseless problems.  
For each BBOB problem, several instances are available, which are obtained from a ``base'' function via translation, rotation and/or scaling transformations~\cite{cocoplat}. 
Each problem instances is a real-valued function $f: [-5,5]^d \to \R$. Problems scale for arbitrary dimensions $d$. In our experiments, we consider six different dimensions, $d \in \{5,10,15,20,25,30\}$, and we focus on the first five instances of each problem (first instance in Sec.~\ref{sec:combi}. In abuse of notation, we shall often identify the functions by their ID $1, \ldots, 24$. 

\textbf{Computation of Feature Values via flacco.} 
For the feature value approximation, we sample for each of the 24 functions $f$ a number $n$ of points $x^{(1)},\ldots,x^{(n)} \in [-5,5]^d$, and we evaluate their function values $f(x^{(1)}),\ldots,f(x^{(n)})$. The set of pairs $\{(x^{(i)},f(x^{(i)})) \mid i=1,..., n\}$ is then fed to the \emph{flacco} package~\cite{flacco2019}, which returns a vector of features. The \emph{flacco} package covers a total number of 343 features~\cite{kerschke_automated_2019}, which are grouped into 17 feature sets. However, some of these features are often omitted in practice because they require adaptive sampling~\cite{BelkhirDSS16,KerschkeT19,morgan_sampling_2014,PitraRH19}, 
while other features have previously been dismissed as non-informative for the \BBOB functions~\cite{ELA_cell,renau_expressiveness_2019}. After removing these sets from our test bed, we are left with six feature sets: 
\textit{dispersion} (disp~\cite{lunacek_dispersion_2006}), 
\textit{information content} (ic~\cite{munoz_exploratory_2015}), 
\textit{nearest better clustering} (nbc~\cite{kerschke_detecting_2015}), 
\textit{meta model} (ela\_meta~\cite{mersmann_exploratory_2011}), 
\textit{$y$-distribution} (ela\_distr~\cite{mersmann_exploratory_2011}), 
and \textit{principal component analysis} (pca~\cite{flacco2019}). 
But even if this selection reduces the number of features to 46, a full enumeration of all subsets for all sizes $c\leq46$ would still be computationally infeasible (since we need to train and test a classification model for each such set). 
We therefore need to reduce the set of eligible features further. 
To this end, we build on the work presented in~\cite{renau_expressiveness_2019}, in which we studied the \textit{expressiveness} of these 46 features. Based on this work we select four features. We add to this selection another six features, one per each of the feature set mentioned above (to ensure a broad diversity of features) and again giving preference to the most expressive ones and to features invariant to \BBOB transformations~\cite{SkvorcEK20}. This leaves us with the following ten features. We indicate in this list by \checkmark and - whether or not a feature is considered invariant under transformation according to~\cite{SkvorcEK20} (first entry) and according to our data (second entry), respectively. Note here that the setting used in~\cite{SkvorcEK20} is slightly different from the instances used in BBOB, mostly due to different ways to handle boundary constraints. The assessment can therefore differ.
\begin{enumerate}
    \item \textbf{disp.ratio\_mean\_02 [\checkmark,\checkmark]} ({\texttt{disp}}) computes the ratio of the pairwise distances of the points having the best 2\% fitness values with the pairwise distances of all points in the design.
    \item \textbf{ela\_distr.skewness [\checkmark,\checkmark]} ({\texttt{skew}}) computes the skewness coefficient of the distribution of the fitness values. This coefficient is a measure of the asymmetry of a distribution around its mean.
    \item \textbf{ela\_meta.lin\_simple.adj\_r2 [\checkmark,\checkmark]} ({\texttt{lr2}}), which computes the adjusted correlation coefficient $R^2$ of a linear model fitted to the data.
    \item \textbf{ela\_meta.lin\_simple.intercept [\checkmark,-]} ({\texttt{int}}), the intercept coefficient of the linear model.
    \item \textbf{ela\_meta.lin\_simple.coef.max [-,-]} ({\texttt{max}}), the largest coefficient of the linear model that is not the intercept coefficient.
    \item \textbf{ela\_meta.quad\_simple.adj\_r2 [\checkmark,\checkmark]} ({\texttt{qr2}}), the adjusted correlation coefficient $R^2$ of a quadratic model fitted to the data. 
    \item \textbf{ic.eps.ratio [-,\checkmark]} ($\varepsilon_{\text{ratio}}$), the half partial information sensitivity.
    \item \textbf{ic.eps.s [-,\checkmark]} ($\varepsilon_s$), the settling sensitivity.
    \item \textbf{nbc.nb\_fitness.cor [\checkmark,\checkmark]} ({\texttt{nbc}}), the correlation between the fitness values of the search points and their indegree in the nearest-better point graph.
    \item \textbf{pca.expl\_var\_PC1.cov\_init [\checkmark,\checkmark]} ({\texttt{pca}}), which measures the importance of the first principal component of a Principal Component Analysis (PCA) over the sample points in the whole search space. 
\end{enumerate}

\textbf{Normalization of Feature Values.}
The value of each feature is normalized between 0 and 1 where 0 (resp. 1) correspond to the smallest (resp. largest) value encountered in the approximated feature values. 
This normalization is performed independently for each dimension, each sample size, and each classifier used in this paper.

\textbf{Sampling Strategy.} Based on an extension of the preliminary experiments reported in~\cite{RenauDDD20} we use a quasi-random distribution to sample the points $x^{(1)}, \ldots, x^{(n)}$ from which the feature values are computed. More precisely, we use Sobol'  sequences~\cite{sobol_distribution_1967}, which we obtain from the Python package \emph{sobol\_seq} (version 0.1.2), 
with randomly chosen initial seeds. 

We sample a total number of 100 independent Sobol' designs, which leaves us with 100 feature value vectors per each function. 
Fig.~\ref{fig:dimension} provides an impression of the distribution of these feature values. Plotted are here approximated values for the \texttt{lr2} feature. 
The comparison shows that the dispersion slightly decreases with the dimension, which is quite surprising in light of the lower density of the points in higher dimensions. We also see that the median values are not stable across dimensions. Some functions (F5 of course, which is correctly identified as a linear function, but also F16, F19, and F20, for example) show a high concentration of feature value approximations, whereas other functions show much larger dispersion within one dimension (e.g., F12, F15, F17, F18) or between different dimensions (F2, F11, F24). 

\begin{figure}[t]
    \centering
    \includegraphics[width=0.9\linewidth]
    {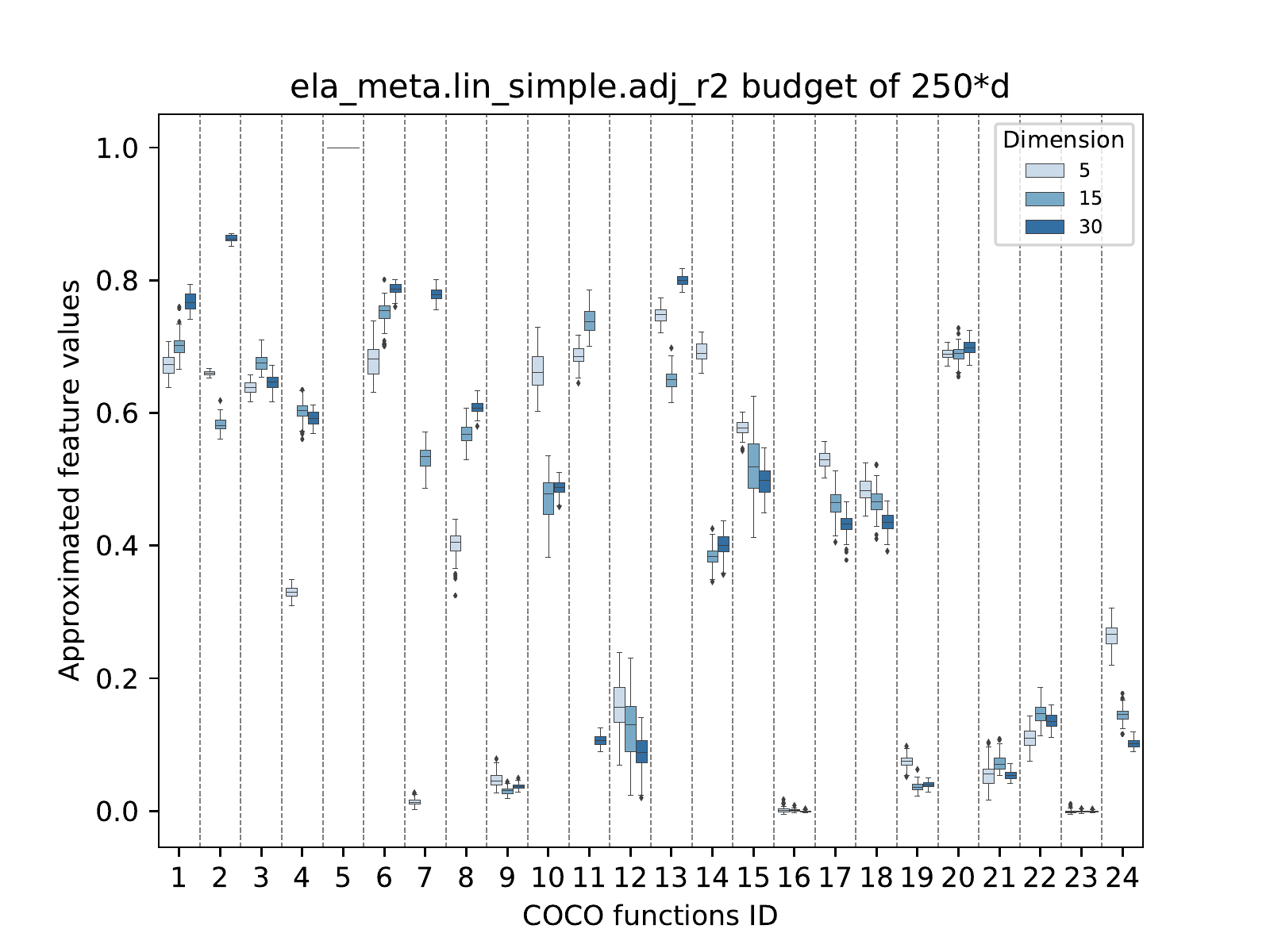}
    \caption{Distribution of the feature values for the \texttt{lr2} 
    feature for different dimensions. Each feature value is computed from $250\times d$ samples and each boxplot represents results of 100 independent feature computations.
    }
    \label{fig:dimension}
\end{figure}

\textbf{Sample Size.} To study the effect of the sample size on the number of features needed to correctly classify the 24 \BBOB functions, we conduct experiments for seven different values of $n$, namely $n \in \{30d, 50d, 100d, 250d, 650d, 800d, 1000d\}$. We note here that a linear scaling of the sample size is the by far most common choice, see, for example,~\cite{BelkhirDSS17,kerschke_low-budget_2016,KerschkeT19}.  

\textbf{Feature Selection.} 
We apply a \emph{wrapper method}, i.e., we actually train a classifier for every considered subset of features. For a given sample size and a given dimension, we train and test all $\binom{10}{c}$ possible subsets of size $c$ starting with $c=1$. If none of these size-$c$ subsets achieves our target accuracy, we move on to the size $c+1$ subsets. As soon as a sufficiently qualified subset has been identified, we continue to evaluate all size-$c$ subsets, but stop the selection process thereafter. This full enumeration of all possible feature combinations for a given size $c$ allows us to investigate the \emph{robustness} of the feature selection. Ideally, we would like to see that the feature sets achieving our 98\% accuracy threshold (this will be introduced below) are stable across the different sample sizes. Robustness with respect to the dimension is much less of a concern to us, since the problem dimension is typically known and can be used for the choosing the feature ensemble that shall be applied to characterize the problem. 

\textbf{Validation Procedure and Target Classification Accuracy.}
In our experiments, we use 80 randomly chosen feature vectors (per function) to train a classification model, and we use the remaining $24 \times 20=480$ feature vectors for testing. For each of these 480 test cases we store the true function ID (i.e., the ID of the function that the feature value originates from) and we store the ID of the function that the classifier matches the feature vector to. From this data we compute the \textit{overall classification accuracy}. 

We repeat this procedure of splitting the set of all feature vectors into 80 training and 20 test instances 20 times; i.e., we repeat 20 times a \textbf{random sub-sampling validation}. We require that the overall classification accuracy \underline{for each} of the 20 validations is \textbf{at least 98\%}. That is, a feature set is eligible if, in each of the 20 random sub-sampling validation runs, it misclassifies at most 10 out of the 480 tested feature vectors. Feature combinations achieving a smaller classification accuracy in one of the validation runs are immediately discarded. 

\textbf{Classification Model.} In the main part of this work, we use a Majority Judgment classifier~\cite{BalinskiL14}. A cross-validation with decision trees and KNN classifiers will be presented in Sec.~\ref{sec:classifier}.

The Majority Judgment classifier works as follows. Let $\Phi=\{\phi_1,\ldots, \phi_k\}$ be the set of features for which we want to know whether it achieves our 98\% target precision requirement. We consider one of the independent subsampling validation runs. That is, for each function we randomly select 80 out of the 100 feature vectors.  
Denote by $\varphi_{i,j,r}$ the $r$-th estimated value for feature $\varphi_i$ for the $j$-th \BBOB function, the set $\{(\varphi_{i,j,r},j) \mid i=1,\ldots, k, j=1, \ldots, 24, r=1,\ldots, 80 \}$ describes the full set of training data. 
From this data we compute for each of the 24 functions $j=1,\ldots, 24$ and for each feature $\phi_i \in \Phi$ the median value 
$$M(i,j):=\mathbb{M}\left( \{ \phi_{i,j,r} \mid r =1,\ldots, 80 \}\right).$$ 
This gives us a set of $24k$ values $M(i,j)$ and concludes the \emph{training step}. 

\begin{table}[t]
\centering
\begin{tabular}{l|llll}
& \multicolumn{4}{c}{\textbf{Function ID (index $j$)}}\\
& \textbf{1 \quad\quad}       &\textbf{ 2 \quad\quad}  & \ldots \quad\quad  & \textbf{24 \quad\quad}    \\
\hline
Feature $\phi_1$ & 0   & \cellcolor{blue!25}0.7 &  \ldots   & 0.7 \\
Feature $\phi_2$ & \cellcolor{blue!25}0.2 & 0.6 &  \ldots   &\cellcolor{blue!25} 0.5 \\
Feature $\phi_3$ & 0.6 & 0.8 & \ldots    & 0.2 \\
\hline
\textbf{Median distance} $D_j$   & \cellcolor{yellow}0.2 & 0.7 & \ldots    & 0.5\\
\hline
\end{tabular}
\caption{Example for the Majority Judgment classification scheme with three features. The values in the table are the distances of the measured feature value $\zeta_i$ to the median feature values $M(i,j)$ of the training set. The median values are reported in the last line. The ID of the function minimizing this median distance $D_j$ is the output of the Majority Judgment classifier. }
\label{tab:MJexample}
\end{table}

In the \emph{testing step} we apply an \emph{approval voting mechanism}~\cite{Brams07} to each of the $480$ test instances. Approval voting mechanisms are single-winner systems where the winner is the most-approved candidate among the voters. From this class of approval voting mechanism we choose \emph{Majority Judgment}~\cite{BalinskiL14} ---a voting techniques which ensures that the winner between three or more candidates has received an absolute majority of the scores given by the voters.

To apply Majority Judgment to our classification task, we do the following. We recall that the task of the classifier is to output, for a given feature vector $\zeta=(\zeta_{i})_{i=1}^k$, the ID of the function that it believes this feature vector to belong to. To this end, it first computes for each of the $k$ features $i$ and for all 24 functions $j$ the absolute distances $d_{i,j}:=|\zeta_{i}-M(i,j)|$. 
Tab.~\ref{tab:MJexample} presents an example for what the distances may look like. 
We then compute for each function the median of these distances, by setting
$
D_j(\zeta):=\mathbb{M}\left(\{d_{i,j} \mid i=1,\ldots, k \}  \right).
$ 
The cells with these median values are highlighted with a blue background in Tab.~\ref{tab:MJexample}, and the values $D_j(\zeta)$ are reported in the last line. The classifier outputs as predicted function ID the value $j$ for which the distance $D_j(\zeta)$ is minimized. This cell is highlighted in yellow background color. 

\textbf{Computation Time.} To give an impression of the computational resources required for our experiments, we report that the computation of the 100 $5$-dimensional feature vectors requires around 6 CPU hours, whereas the computation of the $25$-dimensional feature vectors takes about 1221 CPU hours. Training and testing the classifier takes between 1 second and 3 hours, depending on the setting. In total, we have invested around 432 CPU days for computing the data presented in this work.

\section{Feature Sets Achieving 98\% Classification Accuracy} 
\label{sec:combi}

The portfolios of features for which we obtained the desired 98\% classification accuracy for each of the 20 random sub-sampling validation runs are presented in Tab.~\ref{tab:combis}. For convenience, their sizes are summarized in Tab.~\ref{tab:combisdim}.

\begin{table}[t]
\centering
\begin{tabular}{r|ccccccc}
& \multicolumn{7}{c}{ \textbf{Sample Size}}\\
 \textbf{dimension}  & 30d & 50d & 100d & 250d & 650d & 800d & 1000d \\
\hline
5  & -    & -    & -     & 4     & 4     & -     & 2      \\
10 & -    & -    & -     & 4     & 1     & 2     & 1      \\
15 & -    & -    & 6     & 4     & 2     & 2     & 2      \\
20 & -    & -    & 6     & 2     & 1     & 1     & 2      \\
25 & 1    & 1    & 1     & 1     & 1     & 1     & 1      \\
30 & -    & 6    & 2     & 1     & 1     & 2     & 2     \\
\hline
\end{tabular}
\caption{Feature combination size achieving 98\% classification accuracy in all 20 runs. }
\label{tab:combisdim}
\end{table}

\begin{table}[htp]
  \centering
  \footnotesize{

\begin{tabular}{|c|r|cccccccccc|}
\cline{3-12}\multicolumn{1}{r}{} &       & \multicolumn{10}{c|}{\textbf{Feature}} \\
\hline
\textbf{$d$}
& \multicolumn{1}{c|}{\textbf{$n$}}
&\quad \texttt{int} & \quad \texttt{lr2} & \quad \texttt{qr2} & \quad \texttt{max} & \quad \boldmath{}\textbf{$\epsilon_s$}\unboldmath{} & \quad \boldmath{}\textbf{$\epsilon_{\text{ratio}}$}\unboldmath{} & \quad \texttt{disp} & \quad \texttt{skew} & \quad \texttt{pca} & \quad \texttt{nbc}
\\
\hline
\multirow{7}[2]{*}{\textbf{5}} & \textbf{30$d$} &       &       &       &       &       &       &       &       &       &  \\
      & \textbf{50$d$} &       &       &       &       &       &       &       &       &       &  \\
      & \textbf{100$d$} &       &       &       &       &       &       &       &       &       &  \\
      & \textbf{250$d$} &       & \cellcolor[rgb]{ .988,  .835,  .706}X & \cellcolor[rgb]{ .988,  .835,  .706}X &       &       & \cellcolor[rgb]{ .988,  .835,  .706}X &       &       &       & \cellcolor[rgb]{ .988,  .835,  .706}X \\
      & \textbf{650$d$} &       & \cellcolor[rgb]{ .988,  .835,  .706}X & \cellcolor[rgb]{ .988,  .835,  .706}X &       &       & \cellcolor[rgb]{ .988,  .835,  .706}X &       &       &       & \cellcolor[rgb]{ .988,  .835,  .706}X \\
      & \textbf{800$d$} &       &       &       &       &       &       &       &       &       &  \\
      & \textbf{1000$d$} &       & \cellcolor[rgb]{ .988,  .835,  .706}X &       &       &       & \cellcolor[rgb]{ .988,  .835,  .706}X &       &       &       &  \\
      \hline
\multirow{7}[2]{*}{\textbf{10}} & \textbf{30$d$} &       &       &       &       &       &       &       &       &       &  \\
      & \textbf{50$d$} &       &       &       &       &       &       &       &       &       &  \\
      & \textbf{100$d$} &       &       &       &       &       &       &       &       &       &  \\
      & \textbf{250$d$} &       & \cellcolor[rgb]{ .886,  .42,  .039}\textcolor[rgb]{ 1,  1,  1}{XO} & \cellcolor[rgb]{ .886,  .42,  .039}\textcolor[rgb]{ 1,  1,  1}{XO} &       &       & \cellcolor[rgb]{ .886,  .42,  .039}\textcolor[rgb]{ 1,  1,  1}{XO} &       &       & \cellcolor[rgb]{ .988,  .835,  .706}O & \cellcolor[rgb]{ .988,  .835,  .706}X \\
      & \textbf{650$d$} & \cellcolor[rgb]{ .988,  .835,  .706}X &       &       &       &       &       &       &       &       &  \\
      & \textbf{800$d$} &       &       & \cellcolor[rgb]{ .988,  .835,  .706}X &       &       & \cellcolor[rgb]{ .988,  .835,  .706}X &       &       &       &  \\
      & \textbf{1000$d$} & \cellcolor[rgb]{ .988,  .835,  .706}X &       &       &       &       &       &       &       &       &  \\
      \hline
\multirow{7}[2]{*}{\textbf{15}} & \textbf{30$d$} &       &       &       &       &       &       &       &       &       &  \\
      & \textbf{50$d$} &       &       &       &       &       &       &       &       &       &  \\
      & \textbf{100$d$} & \cellcolor[rgb]{ .988,  .835,  .706}X & \cellcolor[rgb]{ .988,  .835,  .706}X &       & \cellcolor[rgb]{ .988,  .835,  .706}X &       & \cellcolor[rgb]{ .988,  .835,  .706}X & \cellcolor[rgb]{ .988,  .835,  .706}X &       &       & \cellcolor[rgb]{ .988,  .835,  .706}X \\
      & \textbf{250$d$} &       & \cellcolor[rgb]{ .988,  .835,  .706}X &       & \cellcolor[rgb]{ .988,  .835,  .706}X &       & \cellcolor[rgb]{ .988,  .835,  .706}X &       &       & \cellcolor[rgb]{ .988,  .835,  .706}X &  \\
      & \textbf{650$d$} &       & \cellcolor[rgb]{ .988,  .835,  .706}X &       & \cellcolor[rgb]{ .988,  .835,  .706}H & \cellcolor[rgb]{ .988,  .835,  .706}O & \cellcolor[rgb]{ .886,  .42,  .039}\textcolor[rgb]{ 1,  1,  1}{XH} &       &       & \cellcolor[rgb]{ .988,  .835,  .706}O &  \\
      & \textbf{800$d$} &       &       &       &       &       & \cellcolor[rgb]{ .988,  .835,  .706}X &       &       &       & \cellcolor[rgb]{ .988,  .835,  .706}X \\
      & \textbf{1000$d$} &       &       &       &       &       & \cellcolor[rgb]{ .886,  .42,  .039}\textcolor[rgb]{ 1,  1,  1}{XO} & \cellcolor[rgb]{ .988,  .835,  .706}X &       & \cellcolor[rgb]{ .988,  .835,  .706}O &  \\
 \hline
\multirow{7}[2]{*}{\textbf{20}} & \textbf{30$d$} &       &       &       &       &       &       &       &       &       &  \\
      & \textbf{50$d$} &       &       &       &       &       &       &       &       &       &  \\
      & \textbf{100$d$} & \cellcolor[rgb]{ .988,  .835,  .706}X & \cellcolor[rgb]{ .988,  .835,  .706}X &       & \cellcolor[rgb]{ .988,  .835,  .706}X &       & \cellcolor[rgb]{ .988,  .835,  .706}X & \cellcolor[rgb]{ .988,  .835,  .706}X &       &       & \cellcolor[rgb]{ .988,  .835,  .706}X \\
      & \textbf{250$d$} &       &       & \cellcolor[rgb]{ .988,  .835,  .706}X &       &       & \cellcolor[rgb]{ .988,  .835,  .706}X &       &       &       &  \\
      & \textbf{650$d$} & \cellcolor[rgb]{ .988,  .835,  .706}X &       &       &       &       &       &       &       &       &  \\
      & \textbf{800$d$} &       &       &       &       &       & \cellcolor[rgb]{ .988,  .835,  .706}X &       &       &       &  \\
      & \textbf{1000$d$} &       &       & \cellcolor[rgb]{ .988,  .835,  .706}X &       &       & \cellcolor[rgb]{ .886,  .42,  .039}\textcolor[rgb]{ 1,  1,  1}{XO} &       &       & \cellcolor[rgb]{ .988,  .835,  .706}O &  \\
\hline
\multirow{7}[2]{*}{\textbf{25}} & \textbf{30$d$} & \cellcolor[rgb]{ .988,  .835,  .706}X &       &       &       &       &       &       &       &       &  \\
      & \textbf{50$d$} & \cellcolor[rgb]{ .988,  .835,  .706}X &       &       &       &       &       &       &       &       &  \\
      & \textbf{100$d$} & \cellcolor[rgb]{ .988,  .835,  .706}X &       &       &       &       &       &       &       &       &  \\
      & \textbf{250$d$} & \cellcolor[rgb]{ .988,  .835,  .706}X &       &       &       &       &       &       &       &       &  \\
      & \textbf{650$d$} & \cellcolor[rgb]{ .988,  .835,  .706}X &       &       &       &       & \cellcolor[rgb]{ .988,  .835,  .706}O &       &       &       &  \\
      & \textbf{800$d$} & \cellcolor[rgb]{ .988,  .835,  .706}X &       &       &       &       &       &       &       &       &  \\
      & \textbf{1000$d$} & \cellcolor[rgb]{ .988,  .835,  .706}X &       &       &       &       &       &   \cellcolor[rgb]{ .749,  .749,  .749}\textcolor[rgb]{ 1,  1,  1}{\textit{M}}     &       &       &  \\
\hline
\multirow{7}[2]{*}{\textbf{30}} & \textbf{30$d$} &       &       &       &       &       &       &       &       &       &  \\
      & \textbf{50$d$} & \cellcolor[rgb]{ .988,  .835,  .706}X & \cellcolor[rgb]{ .988,  .835,  .706}X &       & \cellcolor[rgb]{ .988,  .835,  .706}X &       & \cellcolor[rgb]{ .988,  .835,  .706}X & \cellcolor[rgb]{ .988,  .835,  .706}X &       & \cellcolor[rgb]{ .988,  .835,  .706}X &  \\
      & \textbf{100$d$} &       & \cellcolor[rgb]{ .988,  .835,  .706}X &       &       &       & \cellcolor[rgb]{ .886,  .42,  .039}\textcolor[rgb]{ 1,  1,  1}{XO} &       &       & \cellcolor[rgb]{ .988,  .835,  .706}O &  \\
      & \textbf{250$d$} & \cellcolor[rgb]{ .988,  .835,  .706}X &       &       &       &       &       &       &       &       &  \\
      & \textbf{650$d$} & \cellcolor[rgb]{ .988,  .835,  .706}X &       &       &       &       &       &       &       &       &  \\
      & \textbf{800$d$} &       &   \cellcolor[rgb]{ .988,  .835,  .706}O    & \cellcolor[rgb]{ .988,  .835,  .706}X &       &       & \cellcolor[rgb]{ .886,  .42,  .039}\textcolor[rgb]{ 1,  1,  1}{XO} &  \cellcolor[rgb]{ .749,  .749,  .749}\textcolor[rgb]{ 1,  1,  1}{\textit{M}}     &       &  &  \\
      & \textbf{1000$d$} &       &   \cellcolor[rgb]{ .988,  .835,  .706}O    & \cellcolor[rgb]{ .988,  .835,  .706}X & \cellcolor[rgb]{ .988,  .835,  .706}H      &       & \cellcolor[rgb]{ .886,  .42,  .039}\textcolor[rgb]{ 1,  1,  1}{XOHV} &  \cellcolor[rgb]{ .749,  .749,  .749}\textcolor[rgb]{ 1,  1,  1}{\textit{M}}     &       &  & \cellcolor[rgb]{ .988,  .835,  .706}V \\
\hline
\multicolumn{11}{c}{}\\
\end{tabular}%

  }
    \caption{Feature combinations achieving the 98\% classification accuracy threshold in all 20 runs. Features with the same symbol (X,O,H,V) belong to the same combination. Results are grouped by dimension $d$ and by the sample size $n$ used to approximate the feature values. Blank rows are for ($d$,$n$) settings for which all $2^{10}$ feature sets failed. M = missing data (due to coronavirus measures in France, we have lost access to cluster and data.) 
  }
  \label{tab:combis}%
\end{table}%

Our first, and most important, finding is that we can actually classify the \BBOB functions with very few features. However, we also see that the existence of such portfolios requires a sufficient sample size. For $d\in \{5,10,15,20\}$, none of the $2^{10}$ possible portfolios based on size-$30d$ and size-$50d$ feature approximations could achieve the 98\% accuracy threshold. 

We also see that, as expected, the size of the minimal portfolio achieving the target precision decreases with increasing sampling size. A few exceptions to this rule exist: 
\begin{itemize}
    \item No combination in $d=5$ with $n=800$ samples achieved the target precision. 
    \item In $d=10$ we see that a single feature, the intercept feature \texttt{int}, suffices to classify with 98\% accuracy when the sampling size is $650d$ and $1000d$. For $800d$, however, this feature does not achieve the threshold. A detailed analysis of the classification accuracy achieved with this feature will be given in Fig.~\ref{fig:intercept}. 
    \item In $d=15$, the $\epsilon_{\text{\text{ratio}}}$ information content feature classifies properly when the sample size equals $n=800d$, but for $n=1000d$, one additional feature is needed to pass the 98\% accuracy threshold. 
    \item In $d=20$ a single feature suffices for $n=650d$ and $n=800d$, but for $n=1,000d$ an additional feature is needed to achieve the target accuracy.
\end{itemize}

Overall, we see that for ten settings a single feature suffices for proper classification. An additional seven cases can be solved by a combination of two features. 
It seems counter-intuitive that in almost all cases the size of the smallest admissible portfolio decreases with increasing dimension. However, as already discussed in the context of Fig.~\ref{fig:dimension}, the dispersion of some feature values \textit{decreases} with increasing dimension -- an effect that is interesting in its own right. Without going into much detail here, we note that this effect is further intensified when using a properly scaled sampling size that maintains the same sampling density across dimensions. 

\textbf{Robustness of the feature combinations with respect to dimension and sample size.} 
Looking at the robustness of the selected combinations over the dimensions and the sample sizes, we observe the following. 

\begin{figure}[t]
    \centering
    \includegraphics[width=0.7\linewidth]{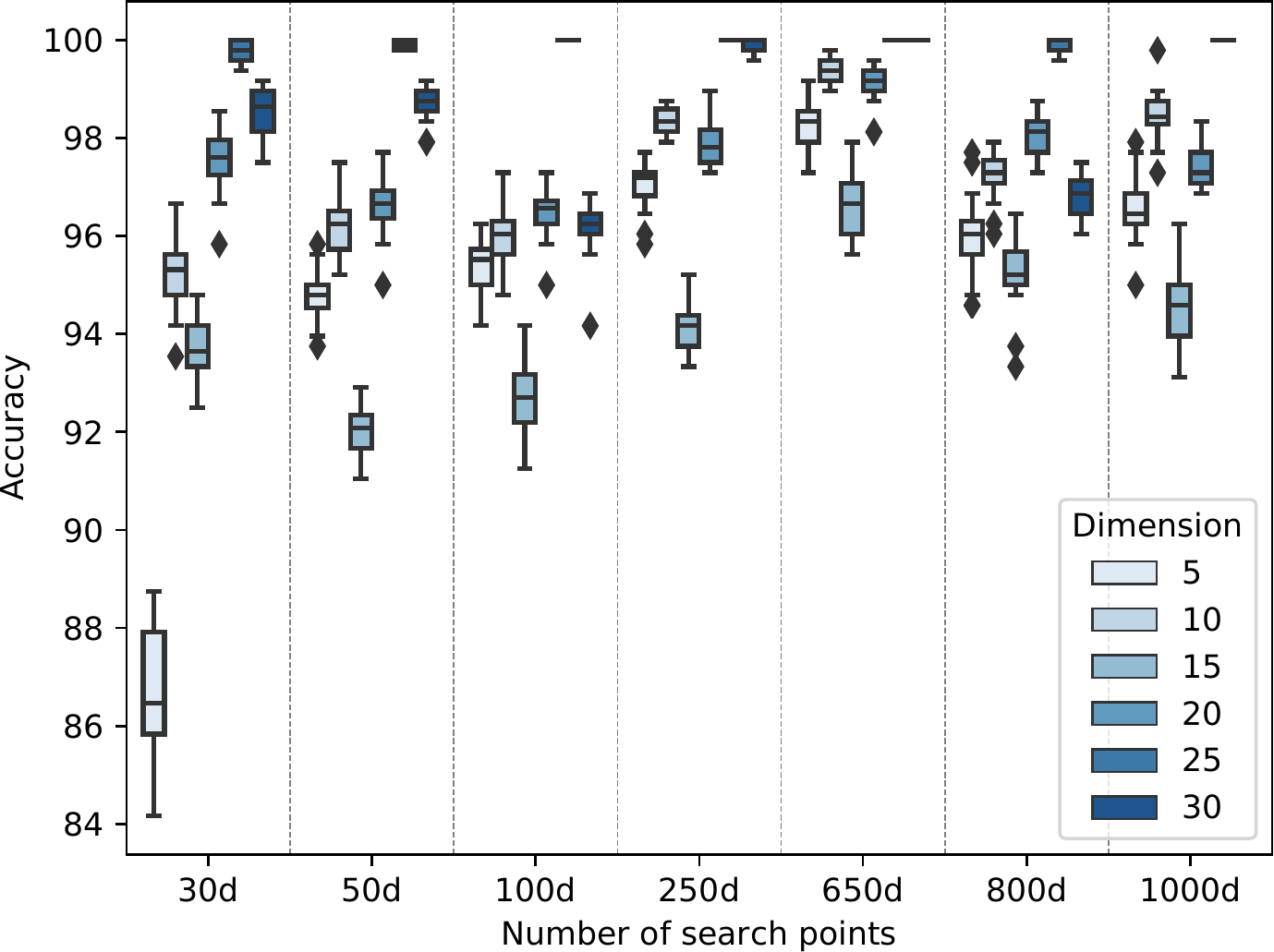}
    \caption{Distributions of intercept feature accuracy by dimension and sample size}
    \label{fig:intercept}
\end{figure}{}

One feature, the intercept feature \texttt{int}, is involved in 15 out of the 28 $(d,n)$ pairs for which a successful feature portfolio could be found. This feature, in contrast, is rarely present in other combinations of size $|c|>1$. To shed more light on its expressive power, we present in Fig.~\ref{fig:intercept} the distributions of the classification accuracy for the various $(d,n)$ combinations. Aggregated over all dimensions and all sample sizes, the median accuracy of the \texttt{int} feature is 96\%. Even if the feature does not always reach our threshold of 98\%, it is worth noting that its performances is almost always above 90\%. Therefore, this feature is very expressive, and this across all tested dimension and sample sizes. 
Another interesting observation from Fig.~\ref{fig:intercept} is that the classification accuracy is not monotonic in the dimension. In all but one case ($n=30d$), the $d=15$ results are worse than those for the other dimensions. As already seen in Tab.~\ref{tab:combis}, for $n=250\times d$ we always have very good classification accuracy. 

The most frequent feature is $\varepsilon_{\text{ratio}}$, which is present in almost all combinations of size $|c|\geq2$.
We count 21 successful combinations of size $|c|\geq2$ and $\varepsilon_{\text{ratio}}$ appears in 20 of these combinations regardless of the dimension and the sample size. In total, it appears in successful portfolios for 17 out of the 28 $(d,n)$ combinations for which a successful subset had been found. The $\varepsilon_{\text{ratio}}$ feature is very useful for our classification task. 

The skewness feature \emph{skew}, in contrast, does not appear in any of the portfolios of the smallest size.

\textbf{Classification Accuracy When Using All flacco Features.} 
We compare the results presented above with the classification accuracy achieved by the Majority Judgment voting scheme using the whole set of 46 features described in Sec.~\ref{sec:setup}. We perform the same sub-sampling validation as above. Interestingly, none of tests performed on the pairs ($d$,$n$) with $n \in \{30d, 50d, 100d, 250d, 650d, 1000d\}$ and $d \in \{5,10,15,20,25,30\}$  met our required target precision of 98\% for each of the 20 runs. We can thus conclude that, in addition to the gain in explainability, the selection of features for supervised-ELA approaches provide better performances, and -- as we shall discuss below -- also come at a much smaller computational cost. 

\section{Robustness with Respect to the Classifier}
\label{sec:classifier}
Having identified feature portfolios that reliably classify the BBOB functions with at least 98\% accuracy when using Majority Judgment (MJ), we now investigate how robust this accuracy is with respect to the choice of the classifier. To this end, we apply the same classification routine as above, but now using \emph{decision trees} (DT) and \emph{$K$ Nearest Neighbors (KNN)} as for classification. We use off-the-shelf implementations from the \emph{scikit learn} Python package~\cite[we use version~0.21.3]{scikit-learn}. Our goal being in investigating robustness, we do not perform any hyper-parameter tuning for these two classifiers. For the KNN classifier we use $K=5$.
For all classifications with a reduced portfolio of features, if multiple combinations are available, only the one marked with \emph{X} in Tab.~\ref{tab:combis} will be used.

Both KNN and decision trees perform as well as our classifier when trained and tested with the small portfolios from Tab.~\ref{tab:combis}, i.e., they both reach at least 98\% classification accuracy in every run except for the decision trees trained with only one feature, for which the accuracy drops to around 62\% in every run. 
Fig.~\ref{fig:boxplot_dimension} summarizes the classification accuracy of the three classifiers for the case that features are based on $n=250d$ samples, for the portfolios described in Tab.~\ref{tab:combis}. Performance is indeed very robust with respect to the classification mechanism.   

\begin{figure}[t]
    \centering
    \includegraphics[width=0.6\linewidth]{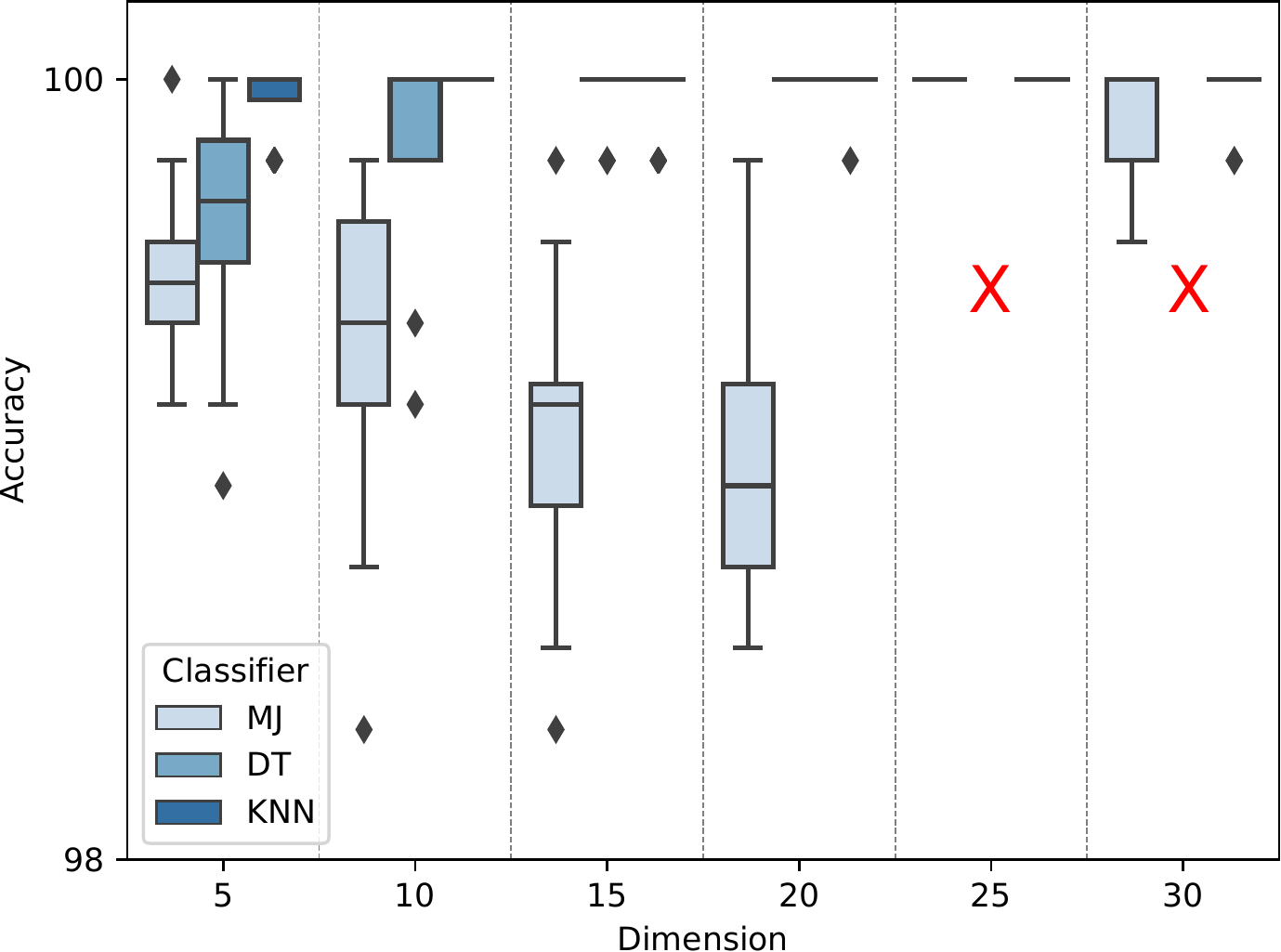}
    \caption{Classification accuracy for the feature portfolios from Tab.~\ref{tab:combis} for budget $250d$. Results are sorted by dimension and classifier and are for 20 random sub-sampling validation runs. Training and testing is done on the first instance of each function only. The \emph{X} corresponds to settings that did not achieve the 98\% threshold.}
    \label{fig:boxplot_dimension}
\end{figure}{}

\textbf{Running Time.} While training and testing were made in around 4 seconds for the DT and for the MJ voting scheme, the KNN classifier needed around 12 seconds to complete the 20 sub-sampling validation runs.

\textbf{Gain over Full Feature Set.} 
We now study how much we gain in terms of computation time when we compute, train, and test the three classifiers (MJ, DT, and KNN) on the selected feature sets only. 

To quantify this gain, we train all three classifiers with the full set of 46 features mentioned in Sec.~\ref{sec:setup}. We first observe that the decision tree classifier has the best performances among the three classifiers in terms of accuracy. It achieves at least 99\% classification accuracy. For KNN, in contrast, performances drops below our 98\% threshold precision on several runs, resulting in a median classification accuracy (over all tests) of around $97\%$. 
The results for KNN align, as already briefly touched upon in Sec.~\ref{sec:combi}, with those obtained using MJ, where none of the tests produced 20 runs in which the threshold was reached.

In terms of computation time, we observe significant differences between the small feature portfolios and the full \emph{flacco} set. As already commented in Sec.~\ref{sec:setup}, the computation of the feature values can be very time-consuming. Reducing the number of features therefore reduces the running time of the feature extraction. However, the savings are even bigger when comparing the cost of training (and testing) the classifiers. 
For decision trees, the execution of the whole classification pipeline takes 3000 times longer than with the small portfolios -- around 3 CPU hours instead of a few seconds. For KNN, the total cost is comparable, also around 3 CPU hours for training and testing the classifiers for the 20 sub-sampling validation runs. For the MJ classifier, the overall running time is only around 35 CPU minutes -- which is still way above the time needed for the small portfolios.

Thus, overall, the reduced portfolios resulted not only in much faster computation times, but achieved also better classification accuracy. 

\section{Robustness with Respect to the Problem Instances} 
The discussion above focused on classifying the first instance of the \BBOB functions, and we now investigate how robust the selection is with respect to different instances of the same problems. Concretely, we investigate classification accuracy when performing the same random sub-sampling validation routine as above to the set of features computed for the first five instances of the \BBOB functions. In this experiment, we keep 80\% of feature values for each instance for training the classifier, and we test on the remaining ones. In a second step we then test transferability, by performing a \emph{leave-one-instance-out (LOIO) cross-validation.} In this setting, the classifiers are trained on four instances of each function and tested on the remaining one. We use the portfolios marked by an X in Tab.~\ref{tab:combis}, and compare to classification accuracy when using all ten features.
In the following, MJ voting is excluded as, by design, it is not suited to work with multiple distributions coming from different instances. Hence, only DT and KNN classifiers will be used in this section.

\begin{figure}[t]
    \centering
    \includegraphics[width=0.6\linewidth]{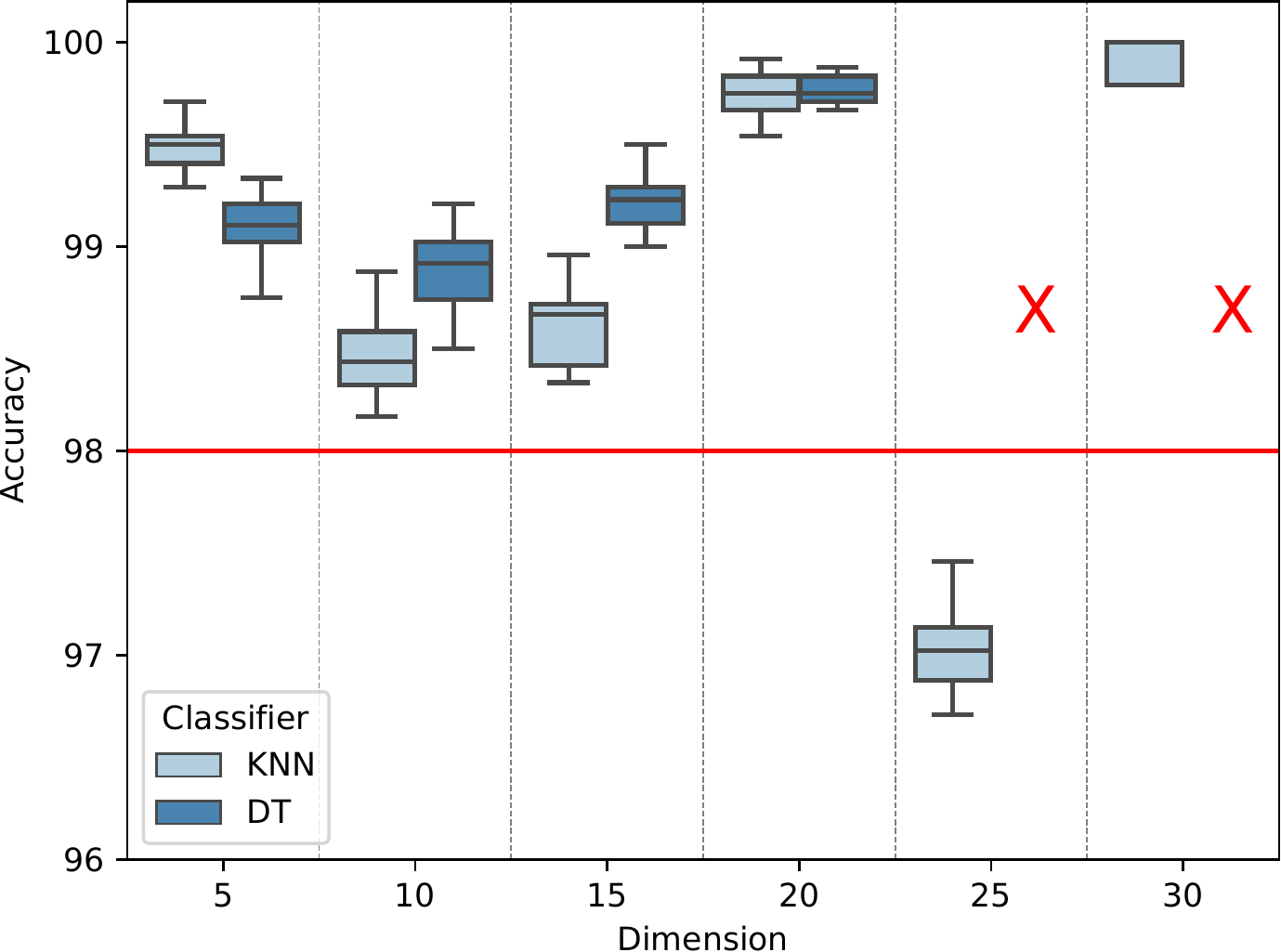}
    \caption{Classification accuracy of DT and KNN classifiers when applied to the first five instances of the 24 BBOB functions. Feature values are computed from $250d$ samples, for the portfolios marked by an X in Tab.~\ref{tab:combis}. Cases with poor performance are marked by a red X.}
    \label{fig:boxplot_dimension_instance}
\end{figure}{}

Fig.~\ref{fig:boxplot_dimension_instance} aggregates the results obtained for the first classification task, where we take feature values from each or the first five instances. 
As in Fig.~\ref{fig:boxplot_dimension}, DT performs badly in $d=25$ and $d=30$, where classification is only based on the intercept feature. For these cases, the median accuracy is 45\% and 62\%, respectively. Since the intercept feature is not invariant to fitness function transformations, the worsened performance is no surprise. In contrast, the median classification accuracy is above 98\% for all portfolios with at least two features. 
We also note that KNN in dimension $d=25$ does not reach our 98\% threshold, but still achieves good performances with an average 97\% accuracy. 

\begin{figure}[t]
     \centering
     \begin{subfigure}[b]{0.45\textwidth}
         \centering
        \includegraphics[width=\textwidth]{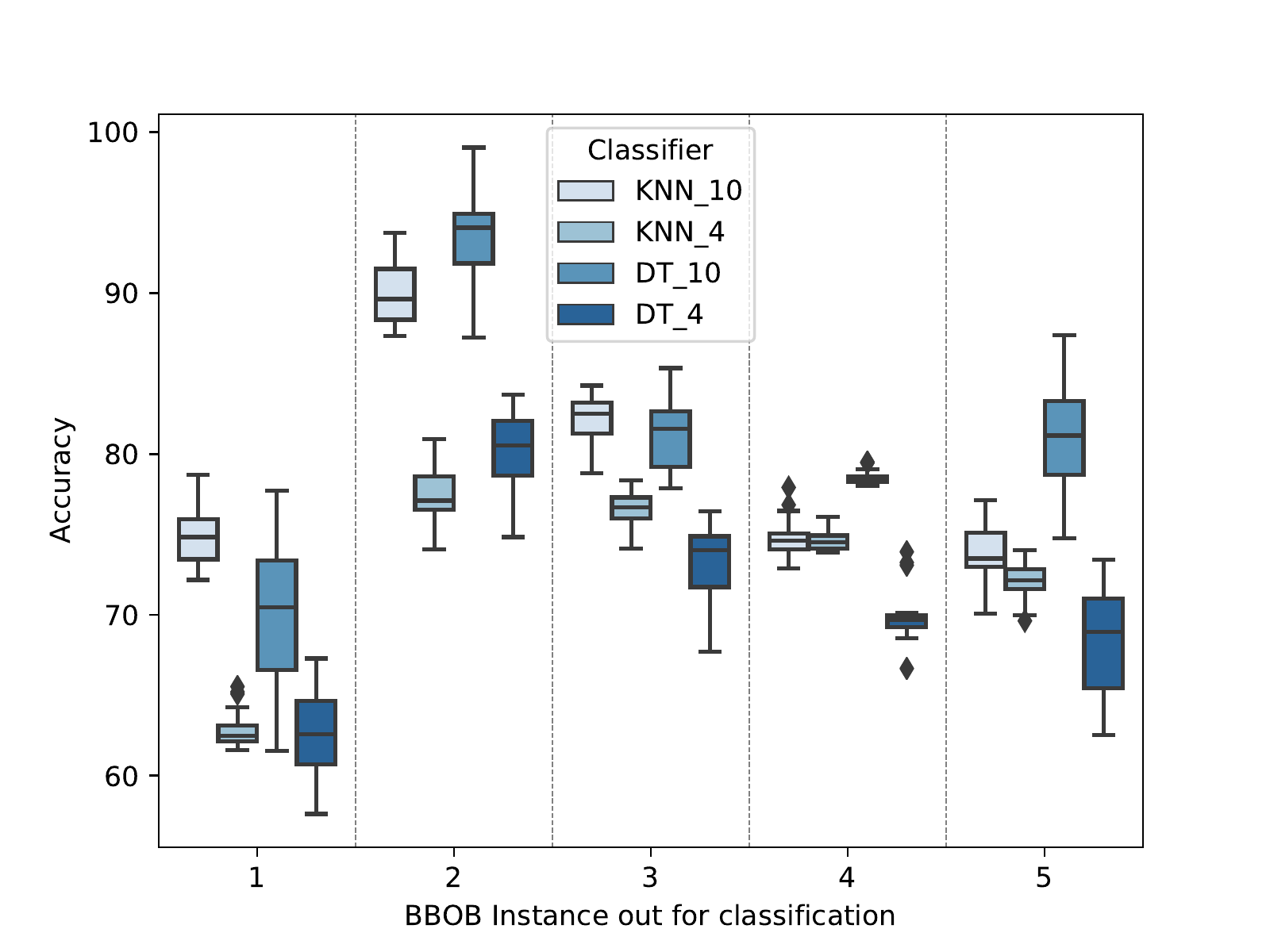}
        \caption{$650d$ samples, $d=5$}
        \label{fig:boxplot_instance1}
     \end{subfigure}
     \hfill
     \begin{subfigure}[b]{0.45\textwidth}
         \centering
        \includegraphics[width=\textwidth]{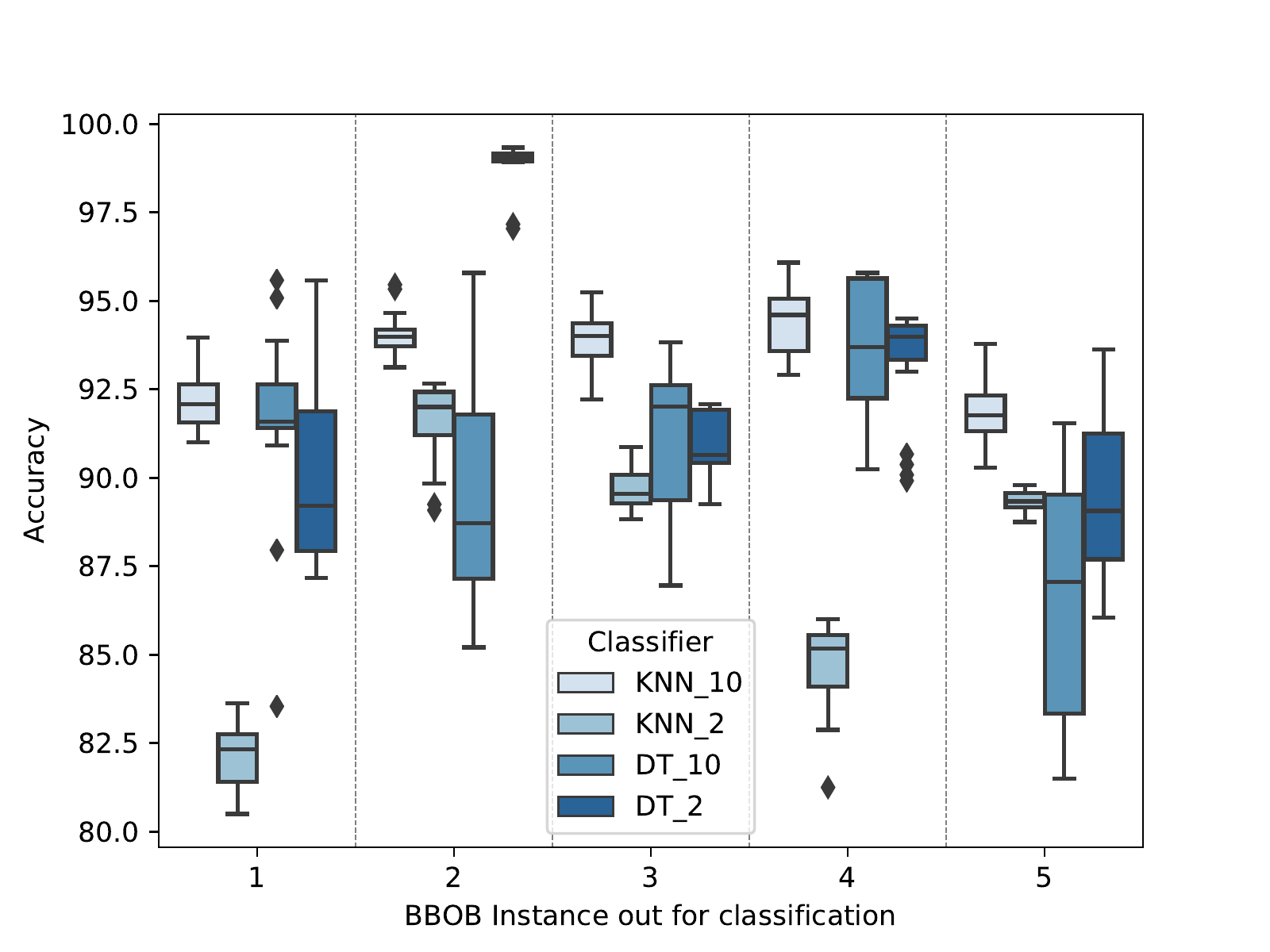}
        \caption{$250d$ samples, $d=20$}
        \label{fig:boxplot_instance2}
     \end{subfigure}
        \caption{Classification accuracy of KNN and DT in the leave-one-instance-out setting. The subscripts $\_2, \_4$, and $\_{10}$ refer to the size of the feature portfolio.}
        \label{fig:boxplots}
\end{figure}

Fig.~\ref{fig:boxplots} presents the classification accuracy achieved by KNN and DT in the LOIO setting. Fig.~\ref{fig:boxplot_instance1} is for features \texttt{lr2}, \texttt{qr2}, $\epsilon_{\text{ratio}}$, and \texttt{nbc} computed from $650d$ samples in $d=5$ and the Fig.~\ref{fig:boxplot_instance2} is for the two features \texttt{qr2} and $\epsilon_{\text{ratio}}$ computed from $250d$ samples in $d=20$. For comparison, we also plot the classification accuracy achieved  when using all ten features listed in Sec.~\ref{sec:setup}. 
For most settings, the accuracy obtained with the set of ten features is better than that achieved for the smaller portfolios. For the $650d$ setting, this is the case for all instances. For the $250d$ setting, DT performs better with the smaller portfolio when instance 1 or instance 3 is left out. The performance loss when using the reduced feature set is particularly drastic for KNN when instance 1 is left out (both cases), when instance 2 is left out ($650d$ case), and when instance 4 is left out ($250d$ case). Interestingly, for DT in the  $650d$ setting, the largest performance losses occur when leaving instance 2 or 5 out. The average loss in classification  accuracy is 5\% and 4\% for KNN in the $650d$ and the $250d$ case, respectively. For DT, the average loss in the $650d$ case is 10\% and the average gain in the $250d$ case is 2\%. 

We conclude that the feature selection is robust when studying different instances, except for those portfolios which consist only of a single feature. For the (arguably more interesting) LOIO setting, however, classification accuracy drops, but non-homogeneously for the different instances. We recommend using the larger feature portfolio in this case. 

\section{Conclusions}
\label{sec:conclusions}

Our ambition to build small feature sets is driven by the desire to obtain models that are (at least to some degree) human-interpretable. While our study certainly has several limitations, as only one test bed is considered, it nevertheless shows that the number of features needed to successfully classify the \BBOB functions is surprisingly low. 
Our main direction for future work is an application of the small feature sets to automated algorithm design tasks. \cite{JankovicD20} shows promising performance of the selected feature portfolio presented in Sec.~\ref{sec:setup} for automated performance regression and per-instance algorithm selection, results that we wish to detail further based on the results presented in Sec.~\ref{sec:combi}.  
Our next important goal will then be to uncover \emph{how} the performance of a given solver depends on the selected features, by taking a closer look at the trained regression models.  With small feature sets, there is reasonable hope that we can identify meaningful correlations.

We are targeting, in the mid-term perspective, classifiers and automated algorithm design techniques that work well on highly constrained problems and which can cope with discontinuities. Extending the results of this work to such problems forms another important next step. 

Other interesting directions for future work include the investigation of new features recently proposed in the literature, (such as, for example, the SOO-based features~\cite{DerbelLVAT19}). We also plan on a closer inspection of the classification results presented above, particularly with respect to the mis-classifications. Functions that are wrongly classified more often than others (a preliminary investigation showed that these mis-classification rates depend on the dimension. In dimensions $d=10$, for example, function 17 is confused with function 21 in 30\% of the tests even when a sample size of $n=10,000$ is used.) Such data can be used, in particular, for training set selection, but also for the \emph{generation} of new problem instances for which the algorithms show some behavior not observable on other instances of the same collection~\cite{Smith-MilesB15,MunozVBS18}. 

\vspace{1ex}
\textbf{Acknowledgments.} 
We thank C\'edric Buron, Claire Laudy, and Bruno Marcon for providing the implementation of the Majority Judgment classifier.
}

\bibliographystyle{splncs04}
\bibliography{renau}

\end{document}